\documentclass[onecolumn,11pt]{article}
\usepackage[top=1in, bottom=1in, left=1in, right=1in]{geometry}
\usepackage{authblk}

\usepackage[numbers,sort&compress]{natbib}
\usepackage{bm}
\usepackage{dsfont}
\usepackage[multiple]{footmisc}
\usepackage{url}

\usepackage[utf8]{inputenc}
\usepackage{amsfonts, amsmath, amssymb, amsthm}
\usepackage{algorithm} 
\usepackage[noend]{algpseudocode}
\usepackage{epsfig, pdfpages}
\usepackage{lineno}

\DeclareMathOperator{\Var}{Var}
\DeclareMathOperator{\Cov}{Cov}
\DeclareMathOperator{\Cor}{Cor}
\newcommand{\Vara}[1]{\Var\left[#1\right]}
\newcommand{\Cova}[2]{\Cov\left(#1, #2\right)}
\newcommand{\Cora}[2]{\Cor\left(#1, #2\right)}
\newcommand{\Expa}[1]{\mathbb{E}\left[#1\right]}

\usepackage{mathtools}
\DeclarePairedDelimiter\ceil{\lceil}{\rceil}

\newtheorem{theorem}{Theorem}[section]

\newtheorem{lemma}[theorem]{Lemma}
\newtheorem{proposition}[theorem]{Proposition}

\newtheorem*{remark}{Remark}

\usepackage{parskip}

\usepackage{subcaption}
\title{Spatiotemporal $k$-means}

\author[1]{Olga Dorabiala}
\author[2]{Devavrat Vivek Dabke}
\author[3]{Jennifer Webster}
\author[1, 4]{J. Nathan Kutz}
\author[1]{Aleksandr Aravkin}
\affil[1]{Department of Applied Mathematics, University of Washington, Seattle, WA, USA}
\affil[2]{Level Ventures, New York, NY, USA}
\affil[3]{Pacific Northwest National Laboratory, Seattle, WA, USA}
\affil[4]{Department of Electrical and Computer Engineering, University of Washington, Seattle, WA, USA}

\newcounter{alphasect}
\def\alphainsection{0}

\let\oldsection=\section\def\section{\ifnum\alphainsection=1\addtocounter{alphasect}{1}\fi\oldsection}%

\renewcommand\thesection{\ifnum\alphainsection=1\Alph{alphasect}\else\arabic{section}\fi}%

\begin{document}
\maketitle
\begin{abstract}
Spatiotemporal data is increasingly available due to emerging sensor and data acquisition technologies that track moving objects.
Spatiotemporal clustering addresses the need to efficiently discover patterns and trends in moving object behavior without human supervision.
One application of interest is the discovery of moving clusters, where clusters have a static identity, but their location and content can change over time.
We propose a two phase spatiotemporal clustering method called {\em spatiotemporal $k$-means} (ST$k$M) that is able to analyze the multi-scale relationships within spatiotemporal data.
By optimizing an objective function that is unified over space and time, the method can track dynamic clusters at both short and long timescales with minimal parameter tuning and no post-processing.
We begin by proposing a theoretical generating model for spatiotemporal data and prove the efficacy of ST$k$M in this setting.
We then evaluate ST$k$M on a recently developed collective animal behavior benchmark dataset and show that ST$k$M outperforms baseline methods in the low-data limit, which is a critical regime of consideration in many emerging applications.
Finally, we showcase how ST$k$M can be extended to more complex machine learning tasks, particularly unsupervised region of interest detection and tracking in videos. 
\end{abstract}
\section{Introduction}
\label{sec:intro}

The widespread use of sensor and data acquisition technologies, including IOT, GPS, RFID, LIDAR, satellite, and cellular networks allows for, among other applications, the continuous monitoring of the positions of moving objects of interest. These technologies create rich spatiotemporal data that is found across many scientific and real-world domains including ecological studies of collective animal behavior, the surveillance of large groups of people for suspicious activity, and traffic management ~\citep{kalnis2005discovering, vieira2009line, jeung2008convoy}. Often, the data collected is large and unlabeled, motivating the development of unsupervised learning methods that can efficiently extract information about object behavior with no human supervision.

Clustering is one of the primary goals of
unsupervised learning.  As such, it has become a critical data mining tool for gaining insight from unlabeled data by grouping objects based on some similarity measure ~\citep{bishop2006pattern,james2013introduction}.
Spatial clustering refers to the analysis of static data with features that describe spatial location,
while spatiotemporal clustering
adds time
as a feature, and algorithms have to consider both the spatial and the temporal neighbors of objects in order to extract useful knowledge ~\citep{birant2007st}.  There are a handful of spatiotemporal clustering classes, some of which track events or object trajectories, but our focus is on moving clusters, where clusters have a static identity, but their location and content can change over time. The moving cluster problem is especially useful in  applications where it is essential to know whether individuals form loose and temporary associations or stable, long-term ones. Applications such as surveillance, transportation, environmental and seismology studies, and mobile data analysis can be considered within this mathematical framework
~\citep{ansari2020spatiotemporal}.

The mathematical formulation for the moving cluster problem is significantly more challenging than for stationary clustering. Most approaches first cluster in space and then aggregate the results over time, as opposed to minimizing a unified objective function. It has been shown that this post-processing approach can lead to erroneous results~\citep{chen2015clustering}. Further, the most popular approaches are built upon density-based clustering methods, which are sensitive to hyperparameter tuning and do not explicitly track cluster centers~\citep{bhattacharjee2021survey}. Finally, while some existing methods operate well on large data, tracking objects over thousands of time steps or more, they exhibit poor performance in the low-data limit, when dynamics are being inferred from either a small number of individuals or over very short windows of time.

We propose a two phase spatiotemporal, unsupervised clustering method, {\em spatiotemporal $k$-means} (ST$k$M), for the moving cluster problem that addresses the aforementioned shortcomings. Phase 1 identifies the loose associations between objects by outputting an assignment for each point at every time step, with the flexibility for points to change clusters between time steps. The clustering objective function provides a unified formulation over space and time and less hyperparameter tuning compared to existing methods. It also provides the functionality to directly track cluster paths without post-processing, allowing ST$k$M is to identify long-term point behavior, even in a dynamic environment. Phase 2 can be optionally applied to the cluster assignment histories from Phase 1 to output stable, long-term associations. In fact, Phase 2 can be applied to any method that outputs an assignment for each point at every time step. The combination of Phase 1 and Phase 2 allows us to analyze the multi-scale relationships within spatiotemporal data. We introduce ST$k$M, theoretically demonstrate the efficacy of the algorithm, evaluate its performance against existing methods on the moving cluster problem, and highlight the use of ST$k$M for more sophisticated machine learning applications.

\section{Related Work}
\label{sec:related}

Spatiotemporal data generally record an object state, an event, or a position in space, over a period of time.
Spatiotemporal clustering can be divided into six classes: event clustering, geo-referenced data item clustering, geo-referenced time-series clustering, trajectory clustering, semantic-based trajectory data-mining, and moving clusters~\citep{ansari2020spatiotemporal}.
Some of the most prominent algorithms, such as ST-DBSCAN
and ST-OPTICS
belong to the second classification~\citep{birant2007st, agrawal2016development}.
Unfortunately, they require four and six input parameters, respectively, heavily influencing the quality of clusters, and they do not provide meaningful cluster centers for analysis. The hyper-parameter tuning of the ten aforementioned parameters becomes critically important for achieving reasonable performance.

The algorithms in this paper are concerned with the final classification scheme: moving clusters.
A moving object is defined by a set of sequences $\langle id, {\bf x}, t \rangle$, where the variable $id$ is the unique identifier for each point, $t$ is time, and ${\bf x}$ is a vector whose components contain the spatial attributes, i.e. the $x$ and $y$ coordinates~\citep{ansari2020spatiotemporal}. Moving clusters have identities (separate from $id$ above) that do not change over time, although their positions and content may change. The prototypical example is a herd of animals, where individual animals can enter or leave the herd at any given time.

Most approaches to the moving cluster problem first cluster in space and then aggregate the results over time. Kalnis et. al proposed running DBSCAN at every time step and defined a moving cluster criteria to associate clusters in successive time steps~\citep{kalnis2005discovering}.
This approach was later extended to the discovery of convoys consisting of at least some points that exist near one another for a minimum number of consecutive time steps~\citep{jeung2008convoy}. Other work identified flocks of objects that stay together for a given window of time ~\citep{vieira2009line}.
The commonality between these approaches was a requirement for moving clusters to exist in some fixed number of consecutive time steps.
In practice, points can split apart and come back together, motivating the proposal of swarms, where a minimum number of objects travel together for at least some proportion of time steps~\citep{li2010swarm}.
Contrastingly, Chen et. al proposed an extension of DBSCAN that incorporates a novel spatiotemporal distance function, where points' distances are their spatial distances from one another if they are temporal neighbors and zero otherwise~\citep{chen2015clustering}.
Their four step process performs even in the presence of noise and missing data, but, like ST-DBSCAN, requires extensive hyper-parameter tuning.

Though substantial work has been done to develop various spatiotemporal clustering techniques, the performance of these methods is rarely compared against one another and implementations are not open source. Recognizing that there was no unified and commonly used experimental dataset and protocol, Cakmak et. al proposed a benchmark for detecting moving clusters in collective animal behavior ~\citep{cakmak2021spatio}. They generate realistic synthetic data with ground truth, and present state-of-the-art baseline methods. Their implemented algorithms extend spatial clustering methods by first assessing whether a data point is density reachable from another data point with respect to both space and time and then employing a splitting and merging process
~\citep{peca2012scalable}.
Additionally, ST$k$M (based on a pre-print of this paper) has been extended to the more abstract metric case involving graphs~\citep{dabke-2023-stgkm, dabke-2023-stgkm-conf, dabke-2024-stgkm-journal}.

\section{Spatiotemporal $k$-means}
\label{sec:stkm}

Drawing inspiration from approaches that define unique spatiotemporal distance metrics, we propose a clustering objective function that provides a unified formulation over space and time and predicts cluster membership for each point at every time step~\citep{izakian2012clustering, chen2015clustering}. We build upon the $k$-means algorithm,
so that cluster centers are explicitly tracked and there are fewer parameters to tune. In a single pass of Phase 1, without post-processing, point membership and dynamic cluster center paths are output. We provide an optional secondary phase that can extract stable, long-term clusters.

\subsection{Phase 1: Loose, Temporary Associations}
The first phase of our method captures loose associations,
with points having the flexibility to change clusters.
We propose a temporal extension of the $k$-means objective function. We focus on $k$-means, because of its simplicity, speed, and scalablity. Also, unlike density-based methods, $k$-means explicitly identifies cluster centers, giving us the ability to directly track the movement of our $k$ clusters. The objective is shown in \eqref{eqn:st-kmeans}.

\begin{multline}
    \min_{{\bf C},  {\bf W}} \sum_{i=1}^N \sum_{j=1}^k \sum_{t=1}^{T} w_{t,j,i} ||{\bf x}_{t,i} - {\bf c}_{t,j} ||^2 + \lambda ||{\bf c}_{t,j} - {\bf c}_{t+1,j}||^2 \hspace{3cm}
    \text{where } {\bf W}_{t,:,i} \in \Delta_1
    \label{eqn:st-kmeans}
\end{multline}

The matrix \noindent ${\bf X}\in \mathbb{R}^{T \times m \times N}$ contains $N$ data points and the matrix ${\bf C}\in \mathbb{R}^{T \times m \times k}$ contains $k$ cluster centers, both of spatial dimension $m$ being tracked over $T$ time steps.
The matrix ${{\bf W} \in \mathbb{R}^{T \times k \times N}}$ contains auxiliary weights
that map the assignment of points to clusters over time.
Instead of restricting the entries of ${\bf W}$ to the discrete set $\{0,1\}$, we allow them
to vary over the closed interval $[0,1]$. This relaxation is used by fuzzy versions of $k$-means and gives the user a way to quantify the extent of each point's membership to a cluster~\citep{nayak2015fuzzy}.
The second term in \eqref{eqn:st-kmeans} associates cluster centers between time frames automatically, as opposed to through post processing, as in~\citep{kalnis2005discovering, jeung2008convoy, vieira2009line, peca2012scalable, cakmak2021spatio}.
Cluster centers maintain their identity because they are penalized for moving apart,
where the parameter $\lambda \in [0,1]$ controls the extent of the penalty.
Objective \eqref{eqn:st-kmeans} requires all points to exist at every time step.
To ensure this criteria is satisfied, data can be divided into time intervals, missing spatial information in an interval can be augmented using interpolation, and intervals with multiple spatial coordinates can be reduced through averaging.

Problem \eqref{eqn:st-kmeans} can be solved using alternating minimization. The centers are updated using the Gauss-Seidel step in \eqref{eqn:cupdate}, and unlike fuzzy versions of $k$-means, which update the weights
with an explicit formula based on points' distances from cluster centers, we use Proximal Alternating Minimization (PAM), as shown in \eqref{eqn:wupdate}~\citep{nayak2015fuzzy}.
PAM gives us control over how quickly weights are updated and can be thought of as a proximal regularization of the Guass-Seidel scheme~\citep{attouch2010proximal}. PAM is guaranteed to converge as long as $d_k > 1.0$. In practice, we set $d_k=1.1$.

\begin{equation}
    {\bf c}_{t,j}^{k+1} = \frac{\sum_{i=1}^n w_{t,j,i} {\bf x}_{t,i} + n \lambda {\bf c}_{t+1, j}}{\sum_{i=1}^n (w_{t,j,i} + \lambda)}
    \label{eqn:cupdate}
\end{equation}

\begin{equation}
    w_{t,j,i}^{k+1} = proj_{\Delta_1}\bigg(w_{t,j,i} - \frac{1}{d_k} ||{\bf x}_{t,i} - {\bf c}_{t,j}||^2 \bigg)
    \label{eqn:wupdate}
\end{equation}

\begin{figure}[t]
    \centering
    \includegraphics[width=.5\textwidth]{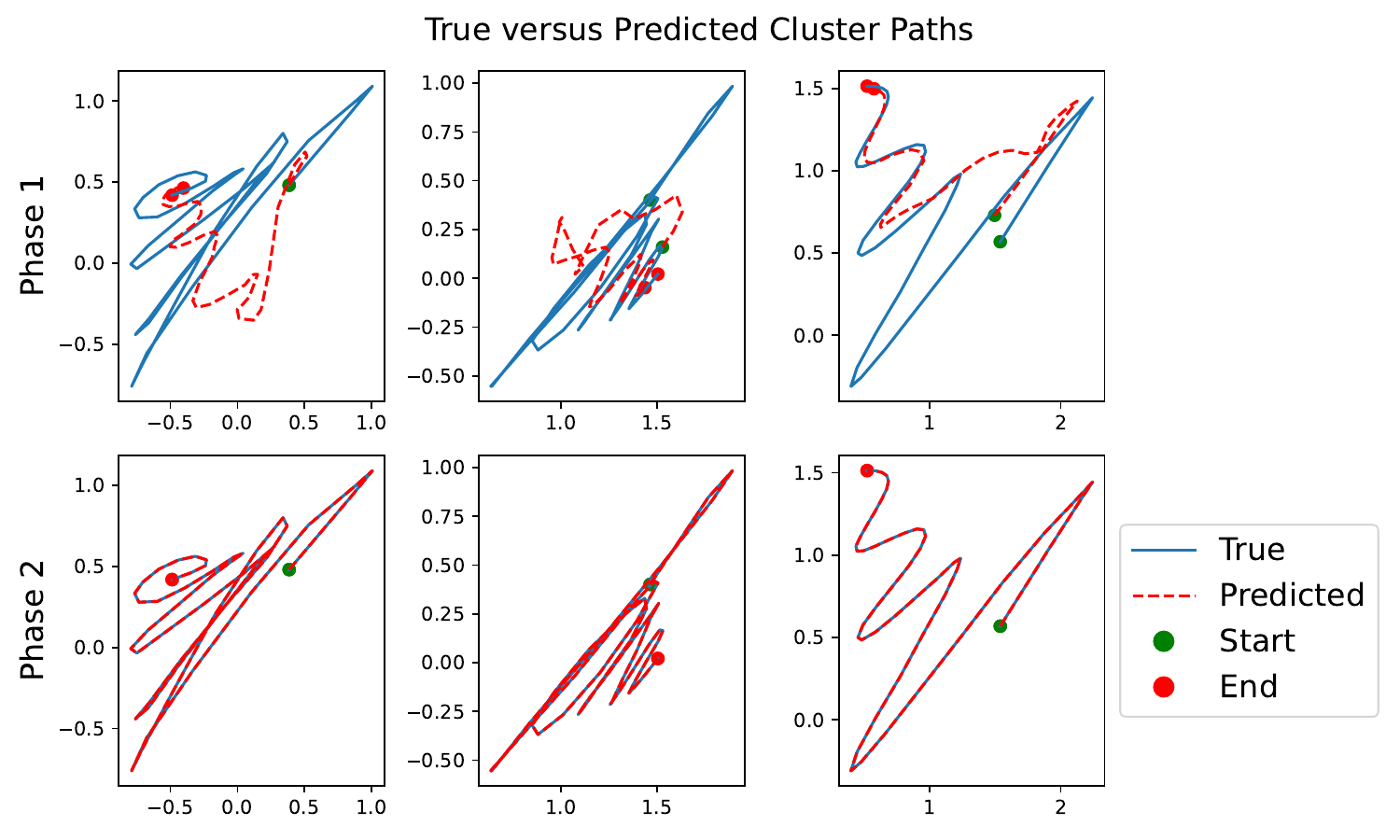}
    \caption{True static versus predicted cluster paths from Phase 1 and Phase 2 of ST$K$M . After Phase 1, ST$k$M identifies general trends of cluster movement, even when allowing points to switch clusters over time. In Phase 2, ST$K$M correctly identifies the true static cluster paths.}
    \label{fig:clusterpaths}
\end{figure}

Since both point membership and clusters are tracked throughout the clustering process
we can directly visualize the paths of dynamic clusters, a feature that without post-processing is unavailable with any existing method.
The top row of Figure \ref{fig:clusterpaths} displays ground truth versus predicted cluster paths using Phase 1 of ST$k$M on a synthetic dataset containing three long-term moving clusters.
Though cluster paths are not identified perfectly because
we do dynamic prediction on clusters with static membership,
ST$k$M is still able to pick up the general trends of cluster movement.
Even in a dynamic environment,
we do not completely lose information about long-term cluster behavior.

\subsection{Phase 2: Stable, Long-term Associations}
Phase 2 of STKM identifies the long-lived associations between data points, and the output
is a single assignment of static clusters containing points that have the most similar spatiotemporal characteristics.
Because Phase 2 uses
dynamic clusters to inform decisions about long-term behavior, the clusters predicted by Phase 2 are more accurate than methods that directly find static clusters.
To run Phase 2 on the output of Phase 1, we first need to extract cluster assignment histories, which we define as
the $\arg\max$ over the rows of ${\bf W}$,
so that the vector ${\bf a}_r\in \mathbb{R}^{T}$ contains the assignment of point $r$ at each time $t$.
We use Hamming distance, denoted as $\mathcal{H}({\bf a}_r, {\bf a}_s)$, to quantify the extent of difference between two vectors ${\bf a}_r, {\bf a}_s$. Then the similarity can be defined as  $\mathrm{sim}({\bf a}_r, {\bf a}_s) = 1 - \frac{\mathcal{H}({\bf a}_r, {\bf a}_s)}{T}$.
We can create a similarity matrix ${\bf A}$, where $A_{r,s}$ contains the similarity between the cluster assignment histories of points $r$ and $s$, and run agglomerative clustering on ${\bf A}$
to output $k$ long-term clusters.

Row two of Figure \ref{fig:clusterpaths}, compares the predicted versus ground truth long-term cluster paths of the moving objects from the previous section, and we observe that Phase 2 identifies the paths perfectly. We can also combine results from Phase 1 and Phase 2 to gain insights about the multi-scale behavior of moving objects e.g. which long-term clusters are most stable, which points switch clusters most often, etc.

\section{Theoretical Analysis}
\label{sec:theory}

\subsection{Overview}

We define the \textit{correlated random walk} model: given a collection of particles with each particle belonging to a unique cluster, the particles are performing random walks that are correlated within a cluster and uncorrelated without.
We would like to analyze how ST$k$M performs on this system. While cluster membership may rightfully change over time in spatiotemporal data, we make the assumption that each particle fits into a unique ``correct" cluster in order to illustrate the utility of ST$k$M.

\subsection{Definitions}

Let \( T \in \mathbb{N} \) be end time for our simulation and \( k \) be the total number of clusters; each cluster has \( n_{i} \) particles (where \( n_{i} \geq 1 \)) and the total number of particles is \( n \triangleq n_{1} + \cdots + n_{k} \), so \( n \geq k \).

Let \( X_{i}^{t} \) represent the position of particle \( i \in [n] \) at time \( t \in \{ 0 \} \cup [T] \) where\footnote{The notation \( [c] \) represents the set \( \{ 1, \ldots, c \} \subset \mathbb{N} \)}, by construction, \( X_{i}^{t} \in \mathbb{R}^d \).
All elements in cluster~\( 1 \) are indexed \( \{ 1, \ldots, n_{1} \} \), those in cluster~\( 2 \) are indexed \( \{ n_{1} + 1, \ldots, n_{1} + n_{2} \} \), and so on; \( a \colon [n] \to [k] \) maps indices to their clusters (see Section~\ref{apn:indexing} for details).

We further construct our \textit{displacements} \( Y_{i}^{t} \).
Let \( W_{i}^{t}, Z_{j}^{t} \stackrel{iid}{\sim} \mathcal{N}(\bm{0}, \mathbb{I}_{d}) \) where\footnote{\( \bm{0} \) is the \( 0 \) element (origin) of \( \mathbb{R}^d \), \( \mathbb{I}_{d} \) is the \( d \times d \) identity matrix} \( i \in [n], j \in [k] \).
We can thus write $Y_{i}^{t} \triangleq \sqrt{q} \cdot W_{i}^{t} + \sqrt{p} \cdot Z_{a(i)}^{t}$
where \( q \triangleq 1 - p \).
These displacements form a set of standard normal vectors that are independent across different timesteps and different clusters, but have correlation \( p \) within a cluster at the same time.
Equivalently, \( Y_{i}^{t} \sim \mathcal{N}\left(\bm{0}, \mathbb{I}_{d}\right) \) with the condition \( \Cora{Y_{i}^{s}}{Y_{j}^{t}} = p \) if \( a(i) = a(j), s = t \) and \( 0 \) otherwise.
Proposition~\ref{prp:equiv} shows equivalence.

\begin{remark}[Correlation is Covariance]
    Since the variance of each \( Y_{i}^{s} \) is \( 1 \), we know that $\Cora{Y_{i}^{s}}{Y_{j}^{t}} = \Cova{Y_{i}^{s}}{Y_{j}^{t}}$
\end{remark}

\subsection{System Dynamics}
To define the dynamics of this system, we let

\begin{equation}
    X_{i}^{t} \triangleq \begin{cases}
        \bm{0}                        & t = 0 \\
        X_{i}^{t - 1} + Y_{i}^{t - 1} & t > 0
    \end{cases}
\end{equation}

\subsection{Key Results}
First, ST$k$M works correctly over time in this context.
In particular, points in the same cluster are more likely to be closer together and ST$k$M exactly optimizes for this case.
We introduce Theorem~\ref{thm:preference} to explain this behavior.
\begin{theorem}
    \label{thm:preference}
    In expectation, intracluster distances are smaller than intercluster distances.
\end{theorem}
\begin{proof}
    Following directly from Lemma~\ref{lma:distribution-of-distance}, observe that the correlation between two points in a cluster is strictly less than the correlation between two points in different clusters.
\end{proof}

Next, we establish a bound on closeness within a cluster.
Theorem~\ref{thm:group-bound} bounds the total distance that a set of points within a cluster can drift.
If a center is chosen within a cluster, then this theorem applies directly where \( q \) is simply\footnote{if the point is not within the cluster, then \( q = 1 \) because \( p = 0 \).} \( 1 - p \).
\begin{theorem}
    \label{thm:group-bound}
    For \( \epsilon > 0 \), the probability that all particles in a cluster are within distance \( D \) of a chosen particle is at least \( 1 - \epsilon \) when \( D = \frac{1}{\epsilon} \cdot c \cdot n_{i} \sqrt{tq} \). The constant \( c \) depends on the ambient dimension, and \( n_{i} \) is the number of particles in the cluster.
\end{theorem}

\begin{proof}
    By Lemma~\ref{lma:distribution-of-distance}, the expected distance of a particle from any particle is \( c \cdot \sqrt{tq} \), where \( c \) is a constant that depends on the ambient dimension\footnote{This lemma applies here because we have selected a cluster, which all have the same correlation with a point, whether or not it is in the cluster.}.
    By Markov's inequality\footnote{Markov's Inequality applies because the distance is non-negative and its expectation is well-defined.}, the probability that this distance is greater than \( D \) is \( \frac{1}{D} \cdot c \cdot \sqrt{tq} \). By the Union Bound, the probability \( \alpha \) that at least one particle is more than distance \( D \) away from a cluster center is at most \( n_{i} \cdot \frac{1}{D} \cdot c \cdot \sqrt{tq} \).
    Let \( D = n_{i} \cdot \frac{1}{\epsilon} \cdot c \cdot \sqrt{tq} \), where \( \epsilon > 0 \).
    By substitution, we see that \( \alpha \leq \epsilon \).
    Finally, we note that by the law of total probability, the probability that no particles are more than distance \( D \) away from any point is simply \( 1 - \epsilon \).
\end{proof}

\section{Experiments}
\label{sec:experiments}

\subsection{Methodology}

To experimentally validate the performance of STG$k$M, we use the benchmark dataset proposed by Cakmak et. al ~\citep{cakmak2021spatio}.
Their benchmark is based on three collective animal behavior models and contains $3,600$ spatiotemporal datasets of sizes ranging from $600$ up to $520,000$, where size is calculated as $T \times n$.
The datasets track static clusters, where points do not change cluster membership over time. During our evaluation, we focus on the datasets that have size between $800$ and $35,000$, of which there are $1,034$. We do so, because we are particularly interested in performance in the low-data domain, where either we have few objects or few time steps from which to infer behavior.

Cakmak et. al measure clustering quality with adjusted mutual information (AMI) score and report execution time for a handful of baseline methods. The methods all output dynamic clusters,
but AMI compares the dynamic cluster assignments against a static ground truth.
To avoid this mismatch, we compare
the ground truth against stable clusters derived from the full assignment histories.
To this end, we use Phase 2 of ST$k$M to extract long-term clusters, not only from Phase 1's output, but also from the baseline methods. Then we report what we refer to as long-term AMI, which compares the predicted versus ground truth static clusters.
We divide our data into groups based on size (e.g. 800-3000, 3000-6000, etc.) and report results as the median and average of
long-term AMI for each range of sizes.
We note that in ~\citep{cakmak2021spatio}, during the cluster merging process, points that cannot be assigned to a cluster are given the same label,
resulting in an erroneous association of unassigned points as a single cluster during the calculation of AMI. In order to avoid this interpretation,
we give them all given unique labels during evaluation.

\subsection{Parameter Selection}
All of the baseline methods have at least four parameters that need to be defined: frame size, frame overlap, $\epsilon_1$, and $\epsilon_2$. These correspond to the number of time steps that belong to a single frame, the number of time steps that frames overlap when associating clusters between frames, and the spatial and temporal distances that define whether a point is density reachable from the current one. All of the methods except for ST-DBSCAN also take as input the true number of clusters $k$. In their experiments, Cakmak et. al arbitrarily fix frame size to be $100$ and frame overlap to be $10$. All of the methods use the default value $\epsilon_1=0.50$, except for ST-DBSCAN, which searches for $\epsilon_1 \in [0.01, 0.05]$. Grid search is used to find the optimal remaining parameters that achieve the highest accuracy measure against the ground truth~\citep{cakmak2021spatio}. In an unsupervised setting, one cannot tune parameters to maximize accuracy based on a ground truth. We argue that the performance of the baseline methods in Cakmak et. al is therefore unrealistic and avoid parameter tuning in our experiments.

In contrast, Phase 1 of our method requires only two parameters: $\lambda$, which controls the extent of the penalty that indirectly discourages points from switching clusters, and $k$, the true number of clusters. The parameter $\lambda$ is confined to the range $[0,1]$, and the meaning of its value is intuitive.
We seek to create a similar, intuitive interpretation of the baseline methods' $\epsilon_2$, the temporal distance a point is density reachable from the current one. We define $\epsilon_2 = \alpha t$, where $t$ is the total number of time steps in the data and $\alpha \in [0,1]$ is some given proportion. This formulation gives us a principled approach to choosing $\epsilon_2$, as opposed to choosing a unique value for each dataset.

Because we know that the ground truth clusters do not allow points to switch clusters, we set both $\lambda$ and $\alpha$ fairly high. We run all of the methods on each set of data with $\lambda, \alpha = [0.60, 0.80, 1.00]$. For the baseline methods, we fix the remaining parameters as follows: frame size $=100$, frame overlap $=10$, $\epsilon_1 = 0.50$ for all of the methods, except for ST-DBSCAN where $\epsilon = 0.05$, and $k$ is set to the true number of clusters. Any other parameters in the baseline methods are set to their default values. Since we run each method three times using different parameters on each dataset, we obtain metrics for 3,102 runs of each method. We then report the aggregate of
long-term AMI for each method on every range of dataset sizes.

\subsection{Results}

\begin{figure}[t]
  \centering
  \includegraphics[width=\linewidth]{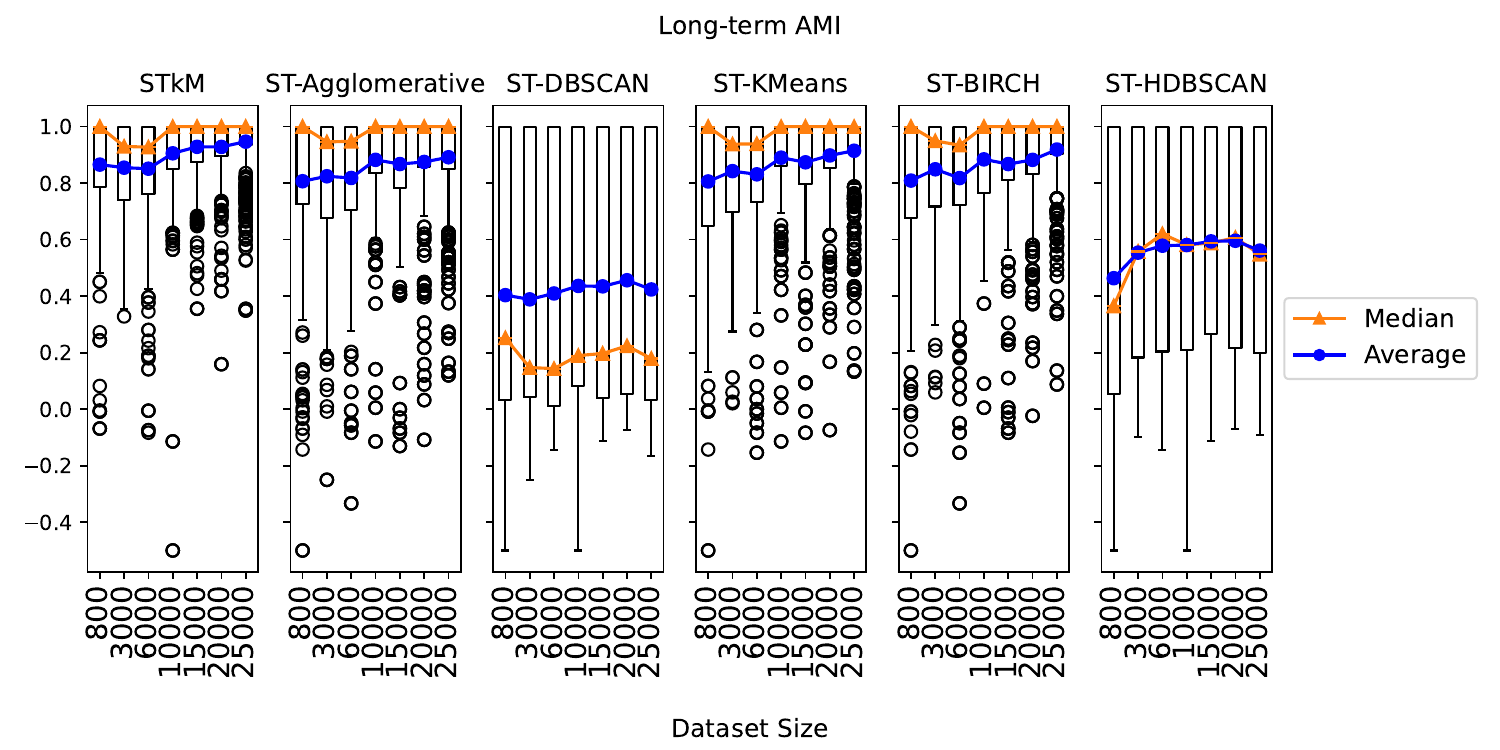}
  \caption{Boxplots of long-term AMI scores for various methods over different dataset sizes. Median scores are shown in orange and average scores in blue. Boxplots for ST$k$M have the top median and average scores, smallest interquartile ranges, shortest tails, and least dispersed outliers, demonstrating that ST$k$M's performance is the best and most consistent. }
  \label{fig:ltamiboxplot}
\end{figure}

Figure \ref{fig:ltamiboxplot} displays the performance of all baseline methods on the benchmark data in terms of long-term AMI.
ST$k$M, ST-Agglomerative, ST-KMeans, and ST-BIRCH score almost identically in terms of their median scores, but ST$k$M maintains the highest averages over all datasets.
As expected, as dataset size increases, more information can be extracted either due to more time steps or more point interactions, and the accuracy of the top methods increases. It is only in datasets under size $10,000$ that we observe median scores noticeably smaller than $1.0$.
Across almost all sizes, the boxplots for ST$k$M in Figure \ref{fig:ltamiboxplot} have the tightest interquartile ranges, the shortest tails, and the most condensed outliers, demonstrating that ST$k$M has the lowest variability and most consistent performance. This result implies that the short-term relationships detected by ST$k$M are the most informative in identifying long-lived point relationships.

\begin{figure}[t]
  \centering
  \begin{subfigure}{.5\textwidth}
    \centering
    \includegraphics[width=.5\linewidth]{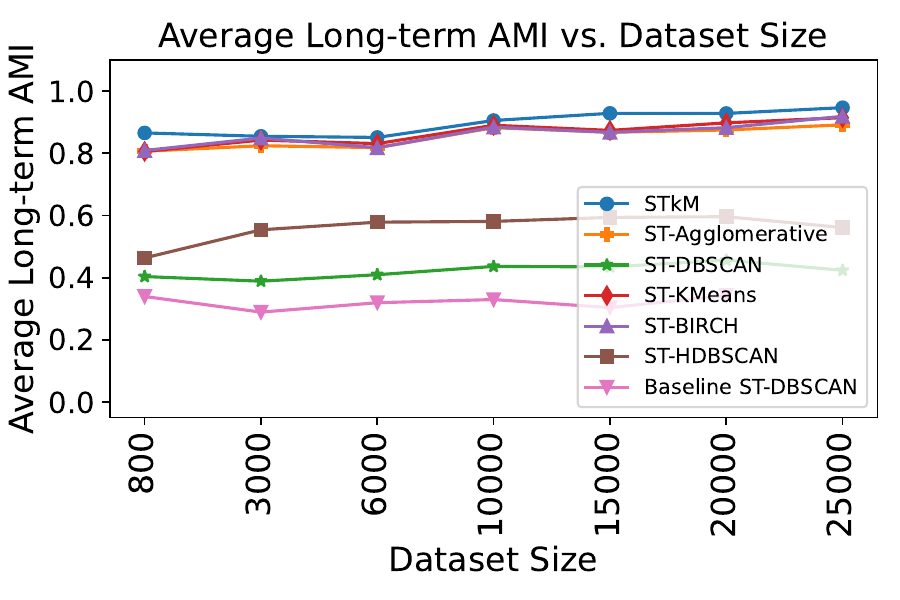}
    \caption{}
    \label{fig:ltamiavg}
  \end{subfigure}%
  \\
  \begin{subfigure}{.5\textwidth}
    \centering
    \includegraphics[width=.5\linewidth]{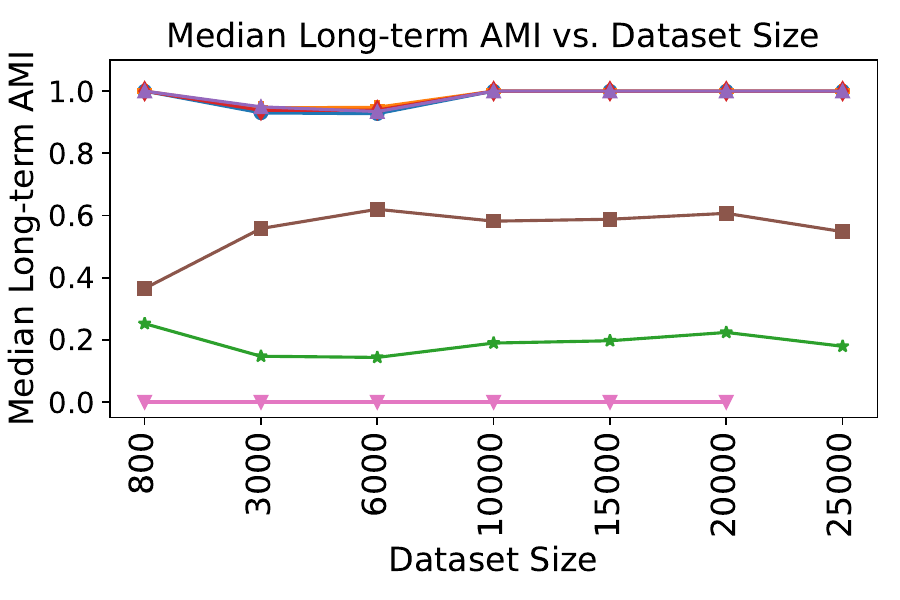}
    \caption{}
    \label{fig:ltamimedian}
  \end{subfigure}
  \caption{(a) Average long-term AMI trendlines. (b) Median long-term AMI trendlines. The top methods perform almost identically in terms of median scores, but ST$k$M achieves the highest average long-term AMIs on datasets of all sizes. All methods that use short-term information to inform long-term predictions perform better than Baseline ST-DBSCAN.}
  \label{fig:ltamiavgmedian}
\end{figure}

\begin{figure}[t]
  \centering
  \includegraphics[width=.5\linewidth]{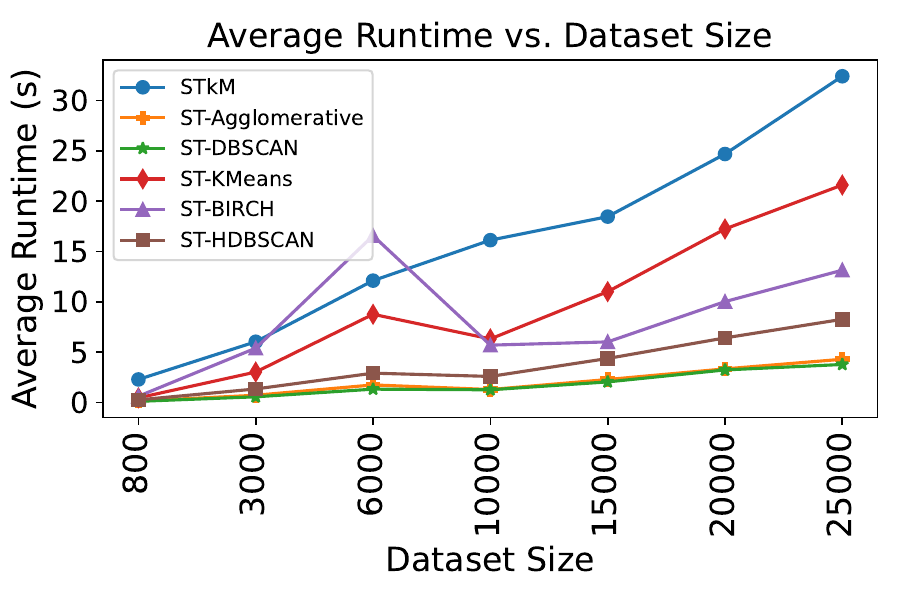}
  \caption{Average runtime versus dataset size. ST$k$M scales poorest in terms of runtime, compared to other methods.}
  \label{fig:rtavg}
\end{figure}

Figure \ref{fig:ltamiavgmedian} provides a closeup of average and median long-term AMI trendlines,
and also includes the results of baseline ST-DBSCAN, which is a popular spatiotemporal clustering method that produces solely static cluster assignments. The remaining methods, which utilize Phase 2 of ST$k$M to generate static cluster assignments, outperform baseline ST-DBSCAN, suggesting that a two phase approach that uses short-term behavior to inform long-term relationships, captures moving object behavior more accurately.
Overall, ST$k$M achieves the highest long-term AMI on $70\%$ of datasets.
Table \ref{tab:overallperformance} shows the long-term AMI scores for each of the tested methods averaged over all 3,102 runs, and we observe that ST$k$M
achieves the highest score.
Although ST$k$M demonstrably outputs more informative moving cluster labels and more accurate long-term cluster labels, the trade-off is it's runtime. ST$k$M runs slowest and scales worst out of all methods tested, as seen in Figure \ref{fig:rtavg}. An improvement could come from decreasing the number of iterations in ST$k$M.

\begin{table*}[!t]
  \caption{Average Long-term AMIs for all methods over all datasets.}
  \begin{center}
    \scalebox{.80}{\begin{tabular}{ |c|c|c|c|c|c|c|}
        \hline
                              & ST$k$M       & ST-Agglomerative & ST-DBSCAN & ST-KMeans & ST-BIRCH & ST-HDBSCAN \\
        \hline
        Average Long-term AMI & \textbf{.90} & .86              & .42       & .87       & .87      & .57        \\
        \hline
      \end{tabular}}
  \end{center}
  \label{tab:overallperformance}
\end{table*}

\subsection{Machine Learning}
Thus far, we have shown ST$k$M’s ability to cluster moving objects of the simplest kind: points traveling in two dimensions. However, moving objects can be much more complex; they can be any evolving, high-dimensional feature vectors. Since ST$k$M sets the benchmark in the two dimensional case, we seek to apply it to more interesting machine learning applications, such as region of interest (ROI) detection and tracking in videos.

Variants of $k$-means have successfully been applied in the context of image segmentation in literature dating all the way back to the 1980s \citep{coleman1979image, pappas1989adaptive}. Over the years, approaches have become more sophisticated, experimenting with pre- and post-processing, ensembling, and the integration of clustering objectives into the functions being optimized by neural networks \citep{dhanachandra2015image, ji2019invariant, kim2020unsupervised}. Extracting ROIs in videos is much more challenging. Most methods use deep learning to extract ROIs on a frame-by-frame basis and aggregate them over time, as in \citep{wu2019sequence}. However, the aggregation is done over consecutive or short time windows, thereby failing to capture a global perspective \citep{lu2019see}. This is where we believe STG$k$M could be of value.

One approach for region of interest tracking in videos, would be to directly apply ST$k$M to the pixels in a video, where each pixel has a feature vector that captures evolving RGB channels. Unfortunately, ST$k$M will not scale well to hundreds of thousands of points, and three features may not be discriminative enough to generate meaningful clusters.
Our approach is to instead use a pre-trained CNN on each video frame to generate “super-pixels” that summarize the important features in each grid box, simultaneously enriching the feature space and diminishing dataset size.
The process is described formally below.

\subsubsection{Lifting an image model to video}
For a given \( w \), length \( l \), and number of channels \( d \), the space of \text{images} is \( \mathbb{R}^{w \times l \times d} \).
In this setting, we will assume that we have an oracle neural network that maps images to some latent space.
In particular, given latent dimension \( n \), we will assume the existence of a neural network \( N \) such that $N \colon \mathbb{R}^{w \times l \times d} \to \mathbb{R}^{w \times l \times n}$.
Given a \textit{movie} \( (x^{t})_{t \in [T]} \) where \( x^{t} \in \mathbb{R}^{w \times l \times d}, T \in \mathbb{N} \), we can construct a set of spatiotemporal points.
Namely, \( p_{i}^{t} \in \mathbb{R}^{n} \) such that $p_{i}^{t} \triangleq N\left(x^{t}\right)_{\sigma(i)}$,
where \( \sigma \) is a bijection for re-indexing such that \( \sigma \colon [w \cdot l] \to [w] \times [l] \), e.g., \( i \mapsto (\ceil{i / w}, i \mod{l}) \).
With this set of points, we use ST\(k\)M to cluster the movie pixels.

\begin{figure}
  \centering
  \includegraphics[width = .4\textwidth]{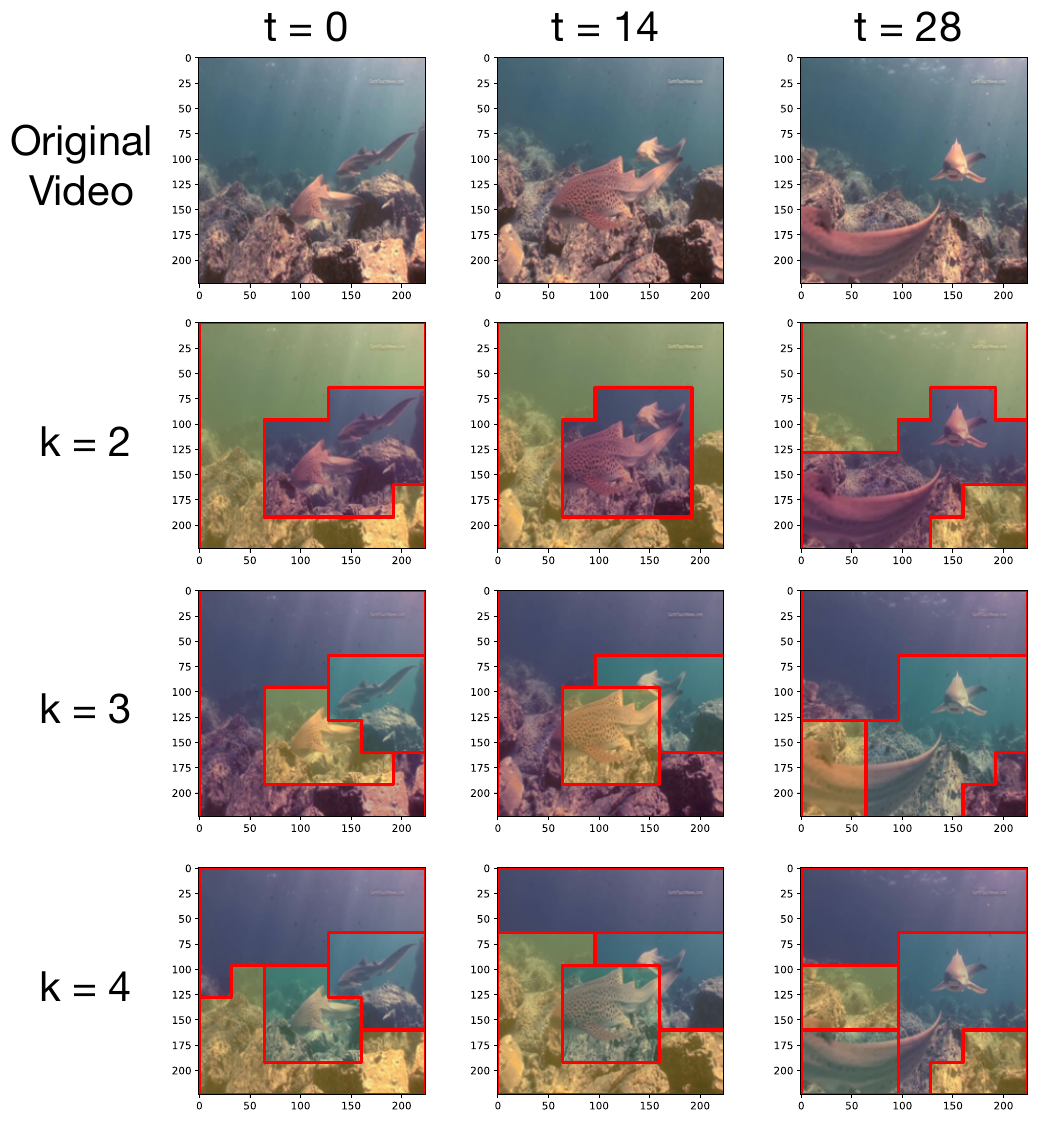}
  \caption{Output of our region of interest detection pipeline on a video of swimming fish using varied values of $k$ in ST$k$M. We achieve background/foreground separation with $k=2$, separate the fish and the water when $k=3$, and cluster coral, water, and fish seperately when $k=4$.}
  \label{fig:fish_figure}
\end{figure}

Figure \ref{fig:fish_figure} shows the output of our proposed pipeline on a video of swimming fish. We run each frame of the video through a pre-trained ResNet 50, with the final layer removed. The output is a 7x7 grid of ``super-pixels” that capture the important features in each grid box. We flatten the grids and run Phase 1 of ST$k$M on the resulting vectors. When $k=2$, we achieve foreground/background separation, assigning the fish and the background to different clusters. When $k=3$, the individual fish are separated from each other and the background. When $k=4$, the clusters correspond to coral, open water, and individual fish. The cluster bounding boxes are not precise; particularly when $k=3$ and $k=4$, small parts of the fish are separated into different clusters. It may be worth experimenting with different CNN backbones or ``super-pixel" granularity, but we leave a principled study and evaluation of ST$k$M for region of interest detection for future work. For now, we emphasize the potential of ST$k$M to be used for this task with no specialized transfer learning, labeled data, or training time.

\section{Conclusion}

We demonstrate that ST$k$M, an unsupuervised two phase spatiotemporal clustering method, is able to capture the multi-scale behavior of moving object data. Phase 1 returns an assignment for each point at every iteration, and provides us the unique ability to directly track cluster centers without any post-processing. This phase minimizes an objective function, that unlike existing methods, is unified in both space and time and requires many fewer parameters to run. Phase 2 can be optionally applied to classify each point into a single long-term cluster. Because Phase 2 infers long-term relationships from short-term ones, Phase 2 results in more accurate static clusters compared to methods that provide exclusively static clusters. The combination of both phases allows us to draw conclusions about the relationships between both points and clusters. 

We demonstrate the competitiveness of ST$k$M against existing spatiotemporal clustering methods on a benchmark dataset proposed by Cakmak et. al ~\citep{cakmak2021spatio}. All algorithms output dynamic clusters, so we use Phase 2 of ST$k$M to translate them into static clusters for comparison against the ground truth. We show that ST$k$M performs best and most consistently in terms of average and median long-term AMI over all datasets, suggesting that the short-term relationships predicted by ST$k$M are more informative than those of the baseline methods. The tradeoff in using ST$k$M is a slower runtime. 

Overall, ST$k$M demonstrably outperforms existing methods on the moving cluster problem. As such, we explore how ST$k$M can be used for more complex machine learning applications and provide evidence that it has the potential to be used as part of an ensemble for region of interest detection in videos. In the future, we intend to explore robust extensions of ST$k$M for handling noise, approaches to estimating the number of clusters $k$, and further study applications of ST$k$M for computer vision and other more complex machine learning tasks. 
In a parallel line of work, we have already extended ST$k$M to the more abstract metric case involving graphs~\citep{dabke-2023-stgkm, dabke-2023-stgkm-conf, dabke-2024-stgkm-journal}.
With ever increasing information from broad applications such as surveillance, transportation, environmental studies, and mobile data analysis, ST$k$M and other related methods are critical for the unsupervised analysis of spatiotemporal data streams. 

\section*{Acknowledgements}
OD and JNK acknowledge support from the National Science Foundation AI Institute in Dynamic Systems
(grant number 2112085)

DD would like to thank Albert Azout and acknowledge support from Level Ventures.

\newpage
{
    \small
    \bibliographystyle{ieeenat_fullname}
    \bibliography{main}
}
\include{sec/10_appendix}
\subsection{Indexing Function}
\label{apn:indexing}
We can assume that we have a surjective assignment function \( a \colon [n] \to [k] \) that assigns each particle to a cluster.
Therefore, our assignment function takes the form
\[ a(i) \triangleq \min \left\{ k' \mathrel{\Big|} i \geq \sum_{j = 1}^{k'} n_{j} \right\} \]
For intuition, note that we are asking for the first cluster such that the total number of points within clusters ``so far'' do not exceed a given input index \( i \).

\subsection{Intermediate Results and Proofs}
We use the following notation:
\begin{itemize}
    \item \( [c] \) represents the set \( \{ 1, \ldots, c \} \subset \mathbb{N} \)
    \item \( \bm{0} \) is the \( 0 \) element (origin) of \( \mathbb{R}^d \)
    \item \( \mathbb{I}_{d} \) is the \( d \times d \) identity matrix
\end{itemize}
\begin{proposition}
    \label{prp:equiv}
    Let \( W_{i}^{t}, Z_{j}^{t} \stackrel{iid}{\sim} \mathcal{N}(\bm{0}, \mathbb{I}_{d}) \) where  \( i \in [n], j \in [k] \).
    Then, $Y_{i}^{t} = \sqrt{q} \cdot W_{i}^{t} + \sqrt{p} \cdot Z_{a(i)}^{t}$ where \( q \triangleq 1 - p \) if and only if \( Y_{i}^{t} \sim \mathcal{N}\left(\bm{0}, \mathbb{I}_{d}\right) \) with the condition \( \Cora{Y_{i}^{s}}{Y_{j}^{t}} = p \) if \( a(i) = a(j), s = t \) and \( 0 \) otherwise.
\end{proposition}
\begin{proof}
    Assume \( Y_{i}^{t} = \sqrt{q} \cdot W_{i}^{t} + \sqrt{p} \cdot Z_{a(i)}^{t} \).
    The sum of two i.i.d. normal distributions is also a normal distribution and the mean and variance of the sum is the sum of the means and variances.
    Therefore, \( Y_{i}^{t} \) has mean \( \sqrt{q} \cdot \bm{0} + \sqrt{p} \cdot \bm{0} = \bm{0} \); it has variance
    \begin{align*}
        \Vara{Y_{i}^{t}} & = \Vara{\sqrt{q} \cdot W_{i}^{t} + \sqrt{p} \cdot Z_{a(i)}^{t}} \\
                         & = q \Vara{W_{i}^{t}} + p \Vara{Z_{a(i)}^{t}}                    \\
                         & = q \mathbb{I}_{d} + p \mathbb{I}_{d}                           \\
                         & = \mathbb{I}_{d}
    \end{align*}
    Thus, \( Y_{i}^{t} \sim \mathcal{N}\left(\bm{0}, \mathbb{I}_{d}\right) \).

    Now, we prove the other direction:
    by construction, the various displacements are uncorrelated at different time steps, so we just need to verify \( \Cora{Y_{i}^{t}}{Y_{j}^{t}} \) as
    \begin{align*}
        \Cora{Y_{i}^{t}}{Y_{j}^{t}} & = \Cora{\sqrt{q} \cdot W_{i}^{t} + \sqrt{p} \cdot Z_{a(i)}^{t}}{\sqrt{q} \cdot W_{j}^{t} + \sqrt{p} \cdot Z_{a(j)}^{t}} \\
                                    & = \Cora{\sqrt{q} \cdot W_{i}^{t}}{\sqrt{q} \cdot W_{j}^{t}}                                                             \\ & \; \; \; + \Cora{\sqrt{q} \cdot W_{i}^{t}}{\sqrt{p} \cdot Z_{a(j)}^{t}} \\
                                    & \; \; \; + \Cora{\sqrt{p} \cdot Z_{a(i)}^{t}}{\sqrt{q} \cdot W_{j}^{t}}                                                 \\
                                    & \; \; \; + \Cora{\sqrt{p} \cdot Z_{a(i)}^{t}}{\sqrt{p} \cdot Z_{a(j)}^{t}}                                              \\
                                    & = 0 + 0 + 0 + \Cora{\sqrt{p} \cdot Z_{a(i)}^{t}}{\sqrt{p} \cdot Z_{a(j)}^{t}}                                           \\
                                    & = p \Cora{Z_{a(i)}^{t}}{Z_{a(j)}^{t}}                                                                                   \\
                                    & = p \mathds{1}_{a(i) = a(j)}
    \end{align*}
\end{proof}

\begin{lemma}[Sequence Correlation is Maintained]
    \[
        \Cora{X_{i}^{t}}{X_{j}^{t}} =
        \begin{cases}
            p & a(i) = a(j)    \\
            0 & a(i) \neq a(j)
        \end{cases}
    \]
\end{lemma}

\begin{proof}

    With \( i, j \) such that \( a(i) \neq a(j) \), \( X_{i}^{t}, X_{j}^{t} \) are sums of i.i.d.\ random variables, so their correlation is \( 0 \).

    If we select \( i, j \) such that \( a(i) = a(j) \), let us compute
    \begin{align*}
        \Cova{X_{i}^{t}}{X_{j}^{t}} & = \Cova{\sum_{r = 0}^{t - 1} Y_{i}^{r}}{\sum_{s = 0}^{t - 1} Y_{j}^{s}}                  \\
                                    & = \sum_{r, s} \Cova{Y_{i}^{r}}{Y_{j}^{s}}                                                \\
                                    & = \sum_{r = s} \Cova{Y_{i}^{r}}{Y_{j}^{s}} + \sum_{r \neq s} \Cova{Y_{i}^{r}}{Y_{j}^{s}} \\
                                    & = \sum_{r = s} p + \sum_{r \neq s} 0                                                     \\
                                    & = t \cdot p
    \end{align*}

    Using the same logic, we can compute
    \begin{align*}
        \Vara{X_{i}^{t}} & = \Cova{\sum_{r = 0}^{t - 1} Y_{i}^{r}}{\sum_{s = 0}^{t - 1} Y_{i}^{s}}                  \\
                         & = \sum_{r, s} \Cova{Y_{i}^{r}}{Y_{i}^{s}}                                                \\
                         & = \sum_{r = s} \Cova{Y_{i}^{r}}{Y_{i}^{s}} + \sum_{r \neq s} \Cova{Y_{i}^{r}}{Y_{i}^{s}} \\
                         & = \sum_{r = s} 1 + \sum_{r \neq s} 0                                                     \\
                         & = t
    \end{align*}

    Putting this together, we can compute
    \begin{align*}
        \Cora{X_{i}^{t}}{X_{j}^{t}} & = \frac{\Cova{X_{i}^{t}}{X_{j}^{t}}}{\sqrt{\Vara{X_{i}^{t}} \cdot \Vara{X_{j}^{t}}}} \\
                                    & = \frac{t \cdot p}{\sqrt{t \cdot t}}                                                 \\
                                    & = p
    \end{align*}

\end{proof}

From here, we study what happens to the distance between two particles over time.
In particular, we want to characterize the distribution of \( \delta(i, j, t) \) where
\[
    \delta(i, j, t) \triangleq \| X_{i}^{t} - X_{j}^{t} \|
\]
From now, we assume \( \| \cdot \| \) is that standard Euclidean distance, but ideally our results and proofs would not depend on the norm selected or would at least work for any \( L^{p} \) norm.

\begin{lemma}[Expectation, Variance of Distance]
    \label{lma:distribution-of-distance}

    Given two particles \( X_{i}^{t} \) and \( X_{j}^{t} \)
    \begin{align}
        \Expa{\delta(i, j, t)} & \in \Theta\left(\sqrt{tqd}\right)  \\
        \Vara{\delta(i, j, t)} & \to tq \tag{as \( d \to \infty \)}
    \end{align}
    In particular, if \( a(i) \neq a(j) \) and thus \( q = 1 \), then
    \begin{align}
        \Expa{\delta(i, j, t)} & \in \Theta\left(\sqrt{td}\right)  \\
        \Vara{\delta(i, j, t)} & \to t \tag{as \( d \to \infty \)}
    \end{align}

    In particular, when \( d = 2\)
    \begin{align*}
        \Expa{\delta(i, j, t)} & \approx 2.5066 \cdot \sqrt{tq} \\
        \Vara{\delta(i, j, t)} & \approx 0.8584 \cdot tq
    \end{align*}
    and when \( d = 3 \) then
    \begin{align*}
        \Expa{\delta(i, j, t)} & \approx 3.1915 \cdot \sqrt{tq} \\
        \Vara{\delta(i, j, t)} & \approx 0.9070 \cdot tq
    \end{align*}

\end{lemma}

\begin{proof}
    Since each component of every \( X^{t} \) is independent, we will begin by analyzing the random vector component-wise.
    We will write \( X_{i, l}^{t} \) to refer to the \( l^{th} \) component of \( X_{i}^{t} \) and related quantities, where \( l \in [d] \).
    First, note that \( X_{i, l}^{t} - X_{j, l}^{t} \) equals
    \[
        \sum_{r = 0}^{t - 1} \left[ \sqrt{q} \cdot W_{i, l}^{r} + \sqrt{p} \cdot Z_{a(i), l}^{r} - \sqrt{q} \cdot W_{j, l}^{r} - \sqrt{p} \cdot Z_{a(j, l)}^{r} \right]
    \]

    In the case that \( a(i) = a(j) \), we see that \( Z_{a(i)}^{t} = Z_{a(j)}^{t} \).
    In the case that \( a(i) \neq a(j) \), we can write \( q = 1, p = 0 \), so in either case, we can write
    \begin{align*}
        X_{i, l}^{t} - X_{j, l}^{t} & = \sum_{r = 0}^{t - 1} \left[ \sqrt{q} \cdot W_{i, l}^{r} - \sqrt{q} \cdot W_{j, l}^{r} \right]  \\
                                    & = \sqrt{q} \sum_{r = 0}^{t - 1} \left[ W_{i, l}^{r} - W_{j, l}^{r} \right]                       \\
                                    & = \sqrt{2tq} \sum_{r = 0}^{t - 1} \frac{1}{\sqrt{2t}} \left[ W_{i, l}^{r} - W_{j, l}^{r} \right]
    \end{align*}
    We can observe that
    \[
        \sum_{r = 0}^{t - 1} \frac{1}{\sqrt{2t}} \left[ W_{i, l}^{r} - W_{j, l}^{r} \right] \sim \mathcal{N}(0, 1)
    \]
    so by letting \( \overline{Z}_{ij, l} \triangleq \sum_{r = 0}^{t - 1} \frac{1}{\sqrt{2t}} \left[ W_{i, l}^{r} - W_{j, l}^{r} \right] \), we can write \( X_{i, l}^{t} - X_{j, l}^{t} = {\sqrt{2tq} \cdot \overline{Z}_{ij, l}} \) and we can now proceed to put everything together.
    Namely, note that
    \begin{align*}
        \left\| X_{i}^{t} - X_{j}^{t} \right\| & = \sqrt{\sum_{l = 1}^{d} \left( X_{i, l}^{t} - X_{j, l}^{t} \right)^2}           \\
                                               & = \sqrt{\sum_{l = 1}^{d} \left( \sqrt{2tq} \cdot \overline{Z}_{ij, l} \right)^2} \\
                                               & = \sqrt{2tq} \sqrt{\sum_{l = 1}^{d} \left(\overline{Z}_{ij, l} \right)^2}
    \end{align*}
    which directly implies that
    \[
        \kappa \cdot \| X_{i}^{t} - X_{j}^{t} \| \sim \mathcal{X}(d)
    \]
    where \( \kappa = 1 / \sqrt{2tq} \).
    The expectation of the \( \mathcal{X} \) distribution is well-known and implies that
    \begin{align*}
        \Expa{\| X_{i}^{t} - X_{j}^{t} \|} & = \sqrt{2tq} \cdot \Expa{\kappa \cdot \| X_{i}^{t} - X_{j}^{t} \|} \\
                                           & = \sqrt{2tq} \cdot \mu_{d}
    \end{align*}
    where \( \mu_{d} \to \sqrt{d - \frac{1}{2}} \in \mathcal{O}(\sqrt{d}) \), but for small values of \( d \), we know that
    \[
        \mu_{d} = \begin{cases}
            \sqrt{2\pi} \frac{ 2^{1 - d} (d - 1)! }{\left(\left( \frac{d}{2} - 1 \right)!\right)^2}                                            & d \text{ even} \\
                                                                                                                                               &                \\
            \sqrt{2} \frac{ \left( \frac{d - 1}{2} \right)! }{ \frac{ 2^{2 - d} \sqrt{\pi} (d - 2)! }{ \left( \frac{d - 1}{2} - 1 \right)! } } & d \text{ odd}
        \end{cases}
    \]

    In particular,
    \begin{align*}
        \mu_{2} & = \sqrt{\frac{\pi}{2}} \approx 1.2533 \\
        \mu_{3} & = \sqrt{\frac{8}{\pi}} \approx 1.5958
    \end{align*}
    so
    \begin{align*}
        \sqrt{2} \cdot \mu_{2} \approx 2.5066 \\
        \sqrt{3} \cdot \mu_{3} \approx 3.1915
    \end{align*}

    Moreover, we can compute that
    \begin{align*}
        \Vara{\| X_{i}^{t} - X_{j}^{t} \|} & = 2tq \Vara{\kappa \cdot \| X_{i}^{t} - X_{j}^{t} \|} \\
                                           & = 2tq (d - \mu_d^2)
    \end{align*}
    which approaches \( tq \) when \( d \) is large, since \( (d - \mu_d^2) \to \frac{1}{2} \)
    and in particular
    \begin{align*}
        2(2 - \mu_{2}^2) & \approx 0.8584 \\
        2(3 - \mu_{3}^2) & \approx 0.9070
    \end{align*}
\end{proof}

\end{document}